\documentclass[a4paper,twoside]{article}

\usepackage{epsfig}
\usepackage{subcaption}
\usepackage{calc}
\usepackage{amssymb}
\usepackage{amstext}
\usepackage{amsmath}
\usepackage{amsthm}
\usepackage{multicol}
\usepackage{pslatex}
\usepackage{apalike}
\usepackage{algorithm2e}
\usepackage[bottom]{footmisc}
\usepackage{xcolor}
\usepackage[dvipsnames]{xcolor}
\usepackage{caption}
\usepackage{array}
\usepackage{tabularx}
\usepackage{booktabs}
\usepackage{url}
\usepackage{subcaption}

\usepackage{SCITEPRESS}     % Please add other packages that you may need BEFORE the SCITEPRESS.sty package.

\title{Eyes on the Grass: \\ Biodiversity-Increasing Robotic Mowing Using Deep Visual Embeddings}

\author{
\authorname{Lars Beckers, Arno Waes, Aaron Van Campenhout and Toon Goedem\'e} \affiliation{EAVISE-PSI research group, Department of Electrical Engineering ESAT, KU Leuven, Belgium}
\email{\{arno.waes, toon.goedeme\}@kuleuven.be}
}


\abstract{
This paper presents a robotic mowing framework that actively enhances garden biodiversity through visual perception and adaptive decision-making. Unlike passive rewilding approaches, the proposed system uses deep feature-space analysis to identify and preserve visually diverse vegetation patches in camera images by selectively deactivating the mower blades. A ResNet50 network pretrained on PlantNet300K provides ecologically meaningful embeddings, from which a global deviation metric estimates biodiversity without species-level supervision. These estimates drive a selective mowing algorithm that dynamically alternates between mowing and conservation behavior. The system was implemented on a modified commercial robotic mower and validated both in a controlled mock-up lawn and on real garden datasets. Results demonstrate a strong correlation between embedding-space dispersion and expert biodiversity assessment, confirming the feasibility of deep visual diversity as a proxy for ecological richness and the effectiveness of the proposed mowing decision approach. Widespread adoption of such systems will turn ecologically worthless, monocultural lawns into vibrant, valuable biotopes that boost urban biodiversity.
}

\begin{document}

\maketitle

\section{INTRODUCTION}

The loss of biodiversity is one of the most pressing ecological challenges of our time. Over the past decades, scientific assessments such as the IPBES Global Report \cite{ipbes2019} and the EU Biodiversity Strategy for 2030 \cite{eubiodiversity2020} have documented steep declines in the abundance and variety of plant and animal species worldwide. Habitat fragmentation, agricultural intensification, and the spread of uniform landscaping practices 
have collectively reduced ecological resilience. Pollinator populations are collapsing \cite{potts2010pollinator}, local flora are being displaced by monocultures, and the structural diversity that underpins ecosystem stability is steadily eroding. According to the European Environment Agency, over 80\% of habitats in Europe are currently in poor conservation status \cite{eea2020}, and urbanization continues to accelerate this trend. 

This decline is not limited to remote natural reserves, it also affects the landscapes closest to where people live. In densely populated regions such as Flanders, Belgium, a surprisingly large share of remaining natural surface consists of private gardens (i.c. 9\% of the land surface, an equal share as the forests in Flanders \cite{Dewaelheyns2014tuinen}). Together, these small, fragmented parcels form an extensive ecological network that, in total surface, rivals or even exceeds that of protected nature reserves. Consequently, increasing biodiversity in gardens could significantly contribute to halting local species loss and restoring ecological continuity in urban and suburban environments. However, their potential is largely untapped: most gardens are dominated by uniform, manicured lawns that have become a cultural ideal 
of neatness and order. 

\begin{figure}[bt]
    \centering
    \includegraphics[width=0.9\linewidth]{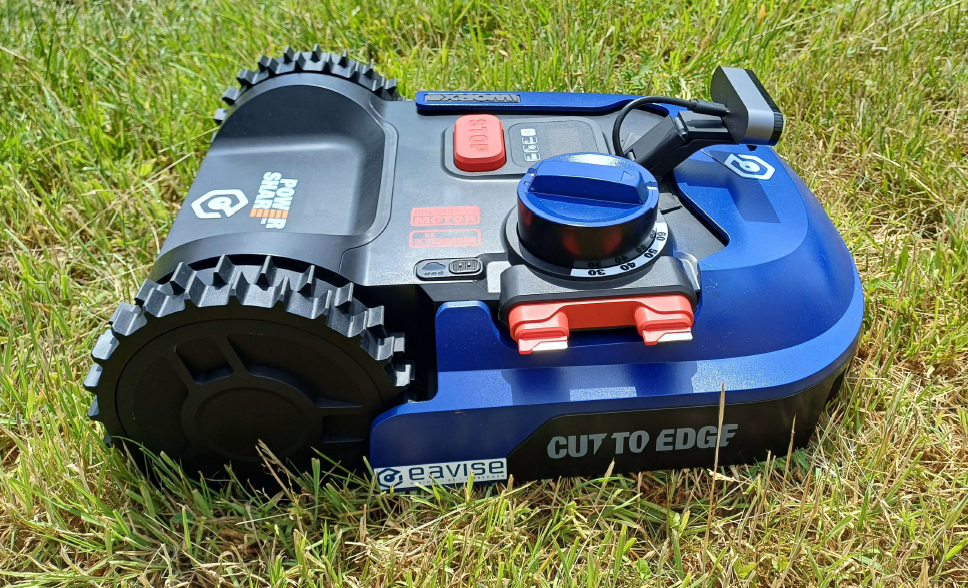}
    \caption{Adapted robotic mower used in the experiments. The camera observes the vegetation in front of the blades; 
    the onboard controller switches the cutting blades on or off based on biodiversity estimation.}
    \label{fig:robot}
\end{figure}

The ideal of the ``perfect lawn'', a uniformly trimmed, bright-green carpet, remains deeply rooted in Western garden culture. 
From an ecological standpoint, such lawns are almost biological deserts: Regular mowing removes flowers before they can seed and prevents insects from feeding or nesting. 
Only a few fast-growing grass species survive the constant disturbance, resulting in a stable but impoverished monoculture. This aesthetic ideal of the ``perfect lawn'' therefore directly conflicts with the need for species diversity in urban ecosystems.

Initiatives such as \textit{``Maai Mei Niet''} (``No Mow May'') have raised public awareness by encouraging temporary pauses in mowing. When mowing is reduced even briefly, wildflowers can bloom, pollinators return, and seed banks are replenished. 
However, once the mower resumes its routine, these gains quickly vanish. The root cause—the constant, homogeneous mowing pattern—remains unaddressed.

A more sophisticated ecological practice known as \emph{sinusoidal mowing} has been implemented in public grasslands and road verges in Flanders and the Netherlands. Rather than mowing in straight, uniform passes, sinusoidal mowing involves mowing along sinusoidal or curved paths at different times and intervals, ensuring that each patch of vegetation is cut at a slightly different moment during the growing season. 
This staggered approach preserves refuge zones for insects and allows seed dispersal between patches, leading to improved species diversity over time. However, sinusoidal mowing is still a manually designed, passive system: its spatial and temporal mowing pattern is predetermined rather than dynamically informed by actual ecological or visual feedback. Our proposed system builds on this idea by making the same principle adaptive and self-regulating 
through continuous visual sensing and local biodiversity estimation.

Robotic lawn mowers have proliferated rapidly in recent years, offering effortless maintenance and perfectly trimmed lawns. Unfortunately, these devices amplify the problem: their programmed regularity enforces uniform grass height across entire plots, ensuring that no plant species can re-establish. Manufacturers are aware of this tension. For instance, Husqvarna's recently introduced \textit{``Rewilding Mode''} allows users to leave a fixed percentage (typically 10\%) of their lawn unmown. While commendable, this is still a static and passive solution. A GPS-defined patch left untouched does not adapt to local ecological variation, and may quickly become dominated by a few opportunistic species.

In contrast, our research introduces an active, data-driven approach to biodiversity-friendly mowing. 
We propose to equip a standard robotic mower with a camera and deep learning algorithms capable of analyzing the vegetation directly in front of it (see fig.~\ref{fig:robot}). Based on the estimated biodiversity contribution of the zone in the captured image w.r.t. the full lawn, the system autonomously decides whether to activate or deactivate its cutting blades. Areas containing plant structures or textures that are underrepresented in the lawn's overall feature space are spared, while overly dominant visual patterns are suppressed by mowing. This local, vision-based decision process aims to create a self-regulating mosaic of vegetation, a garden that remains aesthetically coherent yet biologically rich. 

Our objective is to demonstrate that computer vision and density estimation in deep feature space can serve as a proxy for biodiversity, enabling robotic systems not only to minimize ecological harm but actively to restore diversity at lawn scale.

To validate this concept, we developed a complete research demonstrator of a biodiversity-increasing mowing robot. 
The prototype integrates an onboard camera, a controllable blade system, and an embedded processing unit for real-time vision analysis. Using this platform, we collected image datasets of real and artificial lawns under different mowing regimes and biodiversity levels. These data served as the basis for designing and testing novel computer vision methods for biodiversity estimation directly from RGB images. 

Our contributions in this work are:

\begin{itemize}
    \item We designed and built a functional robot demonstrator that integrates camera-based perception and controllable mowing actuation, enabling proof-of-concept testing of biodiversity-aware mowing strategies.
    
    \item We acquired and curated real-world image datasets from lawns with varying biodiversity, collected under natural lighting and weather conditions, to evaluate the robustness of computer vision approaches in realistic environments.
    
    \item We propose a deep feature space density estimation technique that estimates biodiversity directly from unlabeled lawn imagery. This method uses the embedding representations of pretrained CNNs to quantify visual diversity without explicit species identification.
    
    \item We developed and implemented a selective mowing decision algorithm that uses these density estimates to control mowing locally. The algorithm was validated both in a controlled mock-up lawn experiment, demonstrating the feasibility of adaptive, biodiversity-aware mowing behavior, as well as in a real outdoor lawn setting.
\end{itemize}

\section{RELATED WORK}

Computer vision has increasingly been applied to ecological monitoring, offering scalable and non-invasive tools to quantify biodiversity from images (albeit mostly in remote sensing). While early approaches relied on manual feature extraction or multispectral indices such as NDVI \cite{pettorelli2005ndvi,turner2003remote}, recent advances in deep learning have transformed how visual diversity can be modeled and quantified.

Recent research in biodiversity estimation from imagery can be broadly divided into two categories: \emph{species-level identification} and \emph{representation-level diversity estimation}.  
In the first category, large-scale plant classification datasets such as Pl@ntNet-300K \cite{garcin2021plantnet} and GBIF \cite{gbif2023} have enabled the training of convolutional neural networks (CNNs) for fine-grained recognition of plant species \cite{bhatt2021cnn}. These models achieve high classification accuracy but require extensive labeled data and struggle to generalize to unlabeled, mixed-species vegetation such as lawns or gardens.

In contrast, representation-based methods aim to characterize biodiversity through the structure of the learned feature space itself rather than explicit labels. Inspired by representation learning theory \cite{bengio2013representation}, these approaches use the internal embeddings of pretrained CNNs as a proxy for visual variety. For instance, unsupervised embeddings extracted from ResNet or Vision Transformer models have been used to analyze ecological heterogeneity and habitat variation without species-level annotation \cite{yuan2023foundation}.  
Such deep-feature embeddings capture texture, color, and shape cues that correlate with plant diversity at the patch level, making them suitable for applications where exhaustive labeling is impractical.

Quantifying biodiversity in embedding space requires statistical measures of spread or density. Methods such as k-Nearest Neighbor (kNN) density estimation \cite{zhao2021knn}, Kernel Density Estimation (KDE) \cite{kraml2021xentropy}, and entropy-based dispersion metrics have been explored to assess how densely or sparsely features cluster in representation space. These metrics provide indirect but continuous measures of visual diversity, enabling comparisons across spatial or temporal conditions.  
Mean Shift and DBSCAN clustering algorithms \cite{comaniciu2002meanshift,deng2020dbscan} have also been applied to identify local modes in high-dimensional spaces, which may correspond to dominant vegetation types or homogeneous texture zones within a scene.

Vision-based biodiversity estimation has primarily been investigated at large scales using aerial or satellite imagery \cite{petrou2015remotesensing,aliabad2023urban}. However, few studies have addressed fine-scale, endemically measured biodiversity at the garden or lawn level, where spatial heterogeneity is high and microhabitats play an essential ecological role.  
Most existing techniques remain passive—capturing biodiversity but not acting upon it. Our work builds on this foundation by employing a deep-feature density estimation framework to actively steer a robotic mower toward biodiversity-promoting decisions, thus closing the loop between perception and ecological intervention.

\section{METHODOLOGY}

Our proposed system combines robotic automation and computer vision to enable biodiversity-aware mowing behavior. 
This section describes the mechanical adaptation of the mowing robot, the acquisition of visual data, 
and the computational pipeline that transforms camera images into mowing decisions based on visual diversity estimation.

\subsection{Robot Demonstrator}

To experimentally validate the concept, we adapted a commercial \emph{Landroid M500} robotic mower platform into a research demonstrator (Fig.~\ref{fig:robot}). 
The base robot was equipped with:
\begin{itemize}
    \item a forward-facing RGB camera (Logitec BRIO) mounted on a custom bracket at approximately 20~cm height, 
    providing a $90^{\circ}$ field of view of the grass in front of the blades;
    \item an onboard processing unit (Rock Pi 3A) connected to the camera via USB and powered by the robot's internal battery via a 5V DC/DC convertor;
    \item an electronically switchable relay controlling the mower blades, allowing software-based blade activation or deactivation in real time;
\end{itemize}

Crucially, all perception, inference, decision-making, and blade actuation execute entirely on the onboard module, with no wireless connectivity or offloaded computation. The deployed deep learning models are intentionally lightweight to satisfy strict latency and energy budgets. The additional power draw from the RGB camera and Rock Pi during active inference was measured with a digital multimeter as 
$5.01 \pm 0.32$W. Given the mower’s average operating load of 36W, this overhead shortens its runtime by approximately 12.2\%. The mower therefore returns to its charging station slightly sooner, but mowing performance remains unaffected, demonstrating that fully local ML-based control is feasible even on resource-constrained embedded platforms.

As in the original commercial robot we started from, the robot randomly moves across the lawn. Our setup allows a complete perception–action loop: images are processed onboard, converted to deep embeddings, 
evaluated for biodiversity contribution, and used to control mowing behavior.

\subsection{Proposed Mowing Procedure}
\label{sec:procedure}

\begin{figure}[tb]
    \centering
    \includegraphics[width=0.90\linewidth]{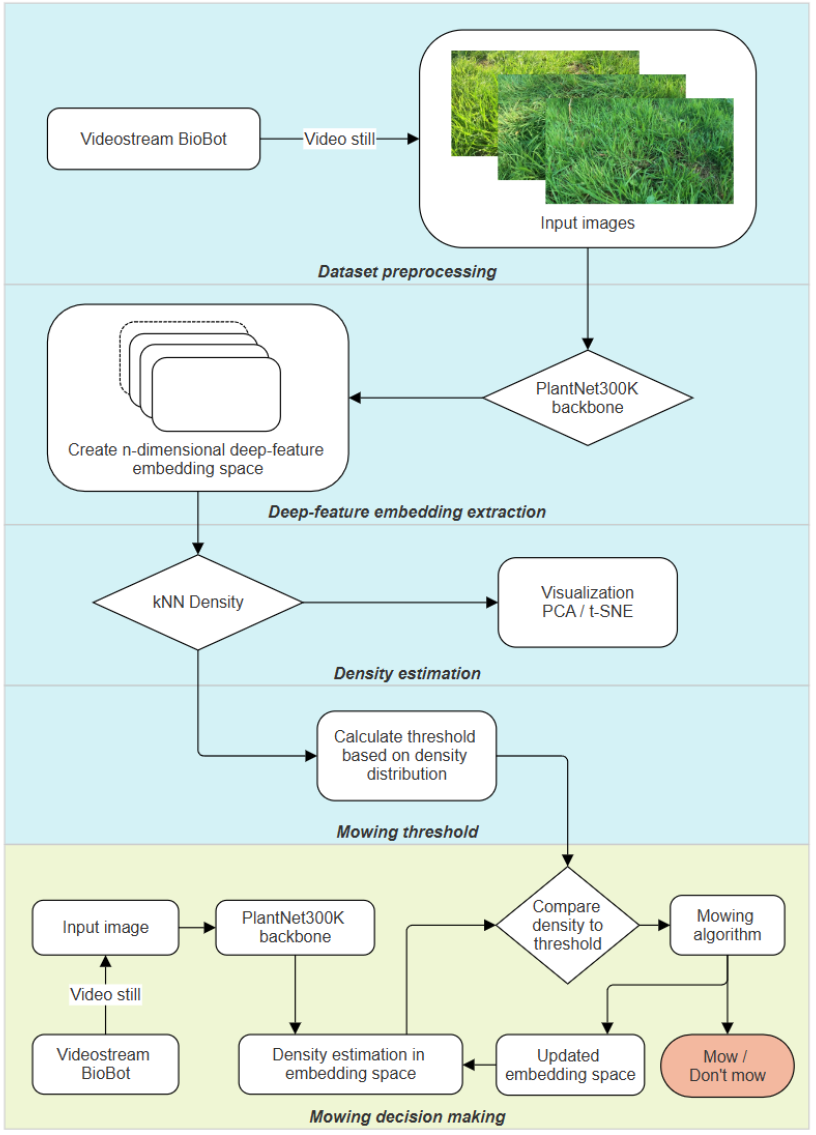}
    \caption{Overview of the mowing procedure}
    \label{fig:procedure}
\end{figure}

The proposed robotic mowing process consists of two stages (Fig.~\ref{fig:procedure}):  
(1) the robot patrols the lawn while collecting images, of which embeddings are computed to construct a biodiversity representation space; (2) during mowing, each new observation is evaluated for local biodiversity importance w.r.t. the garden, and a decision is made to mow or not locally.

\paragraph{Stage~1: Exploratory Patrol and Image Acquisition}

Firstly, the robot performs a full patrol of the lawn without activating its blades. 
The robot follows a random trajectory to capture a representative sample of the vegetation of the lawn. At regular intervals (typically every 0.5--1~m), an RGB image is acquired and stored. These images form the basis for constructing a feature representation of the lawn’s visual diversity (section~\ref{sec:embedding_space}).
Each image $I_i$ is processed by a ResNet50 CNN, pretrained on PlantNet300K, of which the penultimate network layer yields a feature embedding $f_i \in \mathbb{R}^{d}$, with $d=2048$.

The embeddings $\{f_i\}$ collected during the patrol collectively define the biodiversity embedding space of the lawn.  
Intuitively, regions densely populated in this space correspond to frequently observed, visually similar vegetation patches—%
typically dominant grass species—whereas sparse regions indicate rare or unique visual appearances associated with diverse flora.

\paragraph{Stage~2: Biodiversity-Aware Mowing Procedure}

Once the embedding space has been initialized, the robot begins a new mowing trajectory over the same area.  
At each step of the trajectory:
\begin{enumerate}
    \item \textbf{Image capture:} The onboard camera acquires a new image $I_t$ of the vegetation directly ahead.
    \item \textbf{Feature projection:} The image is converted into an embedding $f_t$ using the same CNN backbone 
    and projected into the existing embedding space.
    \item \textbf{Diversity assessment:}  
    The local density $\rho(\bf{f}_t)$ around each embedding $\bf{f}_i$ is estimated using a k-Nearest Neighbor (kNN) approach:
\[
\rho(f_i) = \frac{k}{\sum_{j \in \mathcal{N}_k(f_i)} \|f_i - f_j\|_2 + \epsilon},
\]
where $\mathcal{N}_k(f_i)$ denotes the $k$ nearest neighbors of $f_i$ and $\epsilon$ avoids division by zero.  
This non-parametric density measure acts as a proxy for ecological abundance:  
high density implies dominance; low density implies rarity or potential biodiversity value.
    If the density exceeds a predefined threshold $\tau$, the patch is classified as \textit{overrepresented} 
    (low biodiversity value) and mowing is permitted.  
    If the density is below $\tau$, the area is considered \textit{biodiversity-important}, and the blades remain off.
    \item \textbf{Embedding update:} The new embedding $f_t$ is replacing the older one in the dataset, allowing the representation space to evolve dynamically as the robot explores new microhabitats.
\end{enumerate}

This iterative process creates a feedback loop between perception and action: the embedding space not only represents biodiversity but also evolves through the robot’s interventions.  
Over time, mowing selectively reduces visually homogeneous regions while preserving heterogeneous patches, 
leading to a more diverse vegetation pattern.

%\subsection{General Assumption}
%A pretrained plant-classification model encodes biodiversity-relevant information in its embedding space. 
%Density in this space reflects species abundance; higher spread implies greater biodiversity.

\section{EXPERIMENTS \& RESULTS}

% \subsection{Implementation Details}

% The algorithm was implemented in Python using PyTorch for feature extraction and NumPy for kNN computations.  
% The density threshold $\tau$ and neighborhood size $k$ were determined empirically based on pilot data from 
% lawns of varying biodiversity.  
% All computations were performed onboard in near real time (approx.\ 2~fps).  
% During experiments, mowing decisions were logged and visualized as overlaid color maps, 
% with mowing regions indicated in red and protected regions in green.

\subsection{Datasets}

\begin{figure*}[t]
    \centering

    \begin{subfigure}[t]{0.22\linewidth}
        \includegraphics[width=\linewidth]{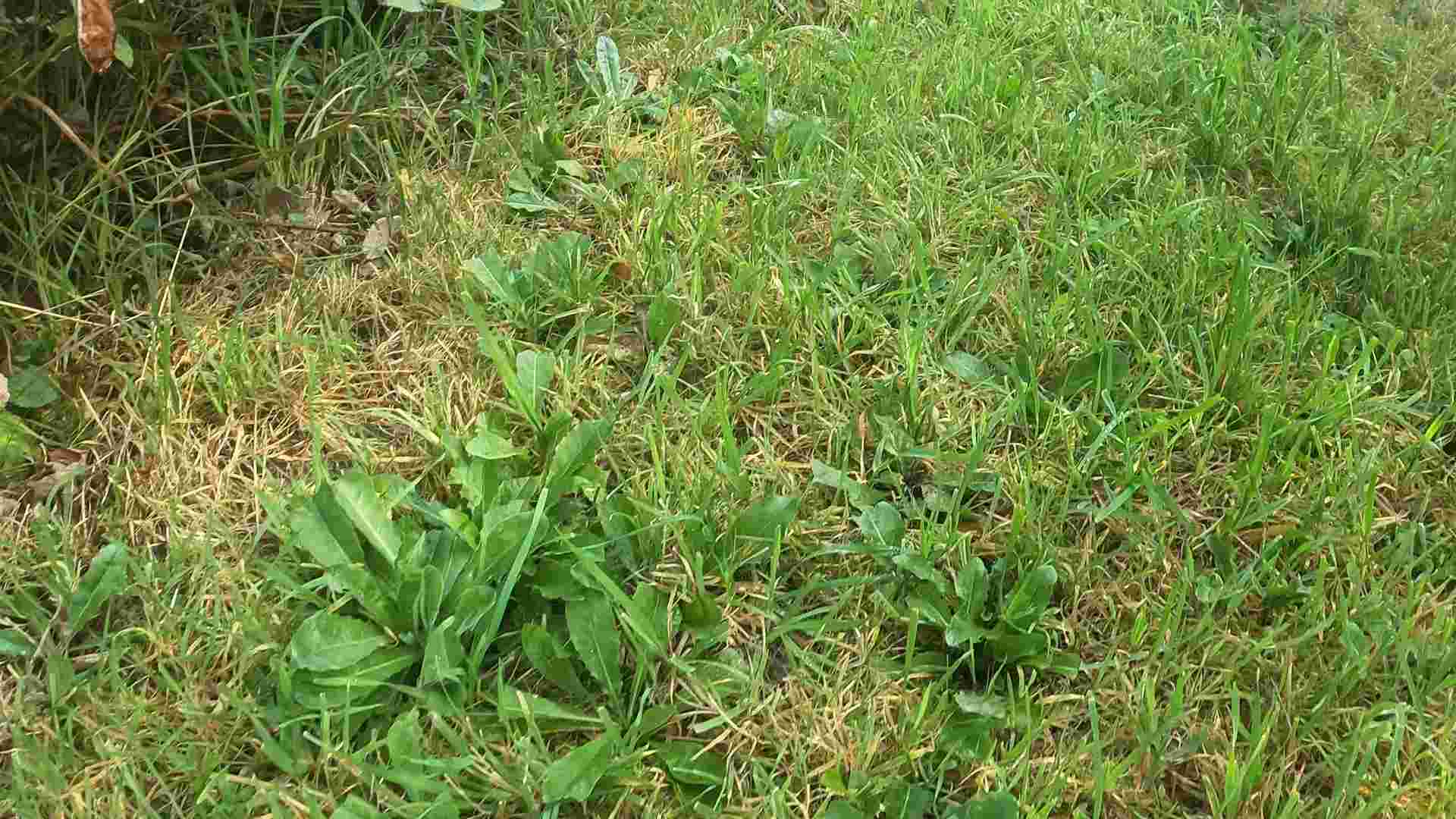}
        \caption{$\color{red} \bullet$ D1}
    \end{subfigure}
        \hfill
    \begin{subfigure}[t]{0.22\linewidth}
        \includegraphics[width=\linewidth]{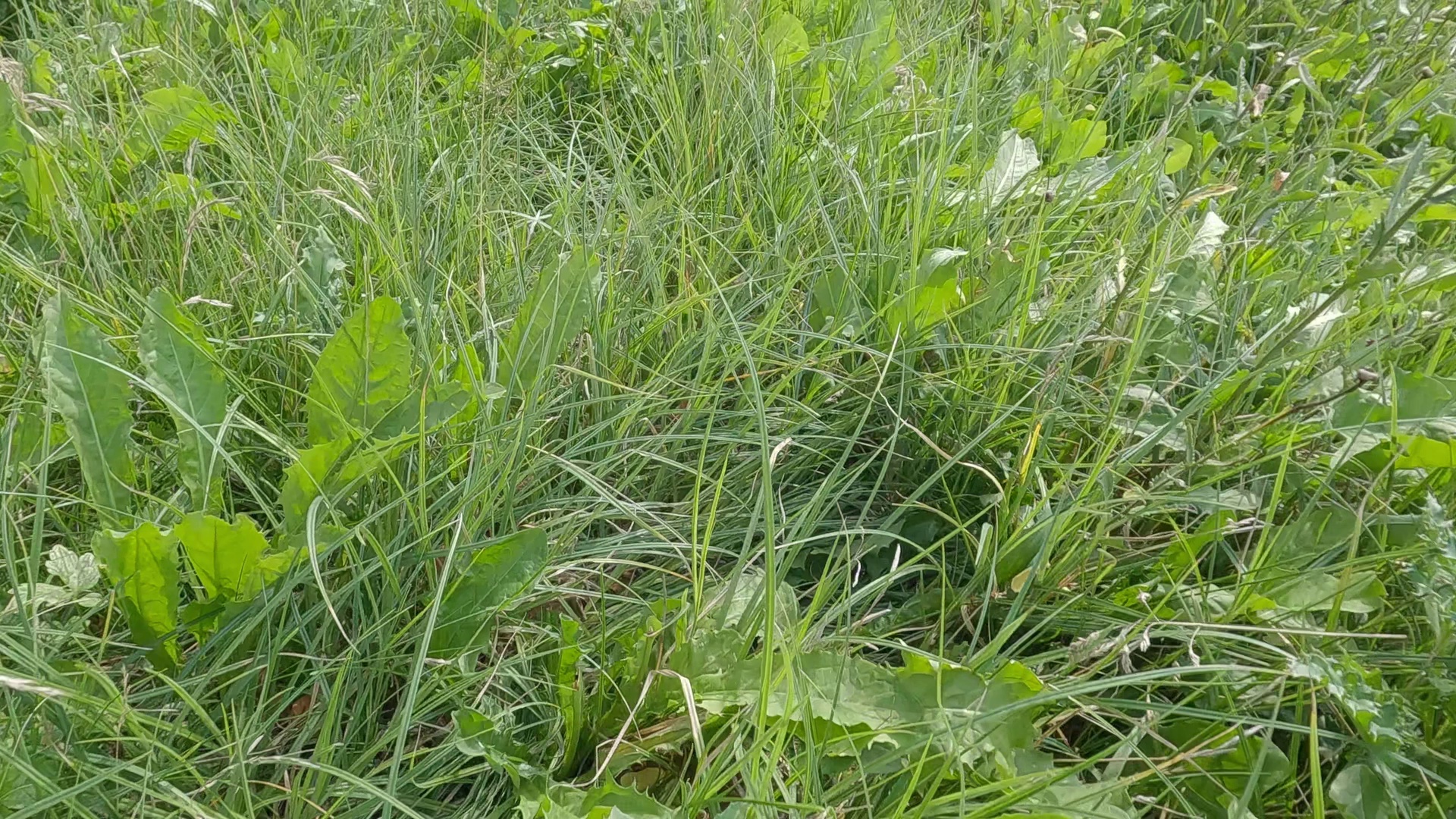}
        \caption{$\color{OliveGreen} \bullet$ D2 }
    \end{subfigure}
    \hfill
    \begin{subfigure}[t]{0.22\linewidth}
        \includegraphics[width=\linewidth]{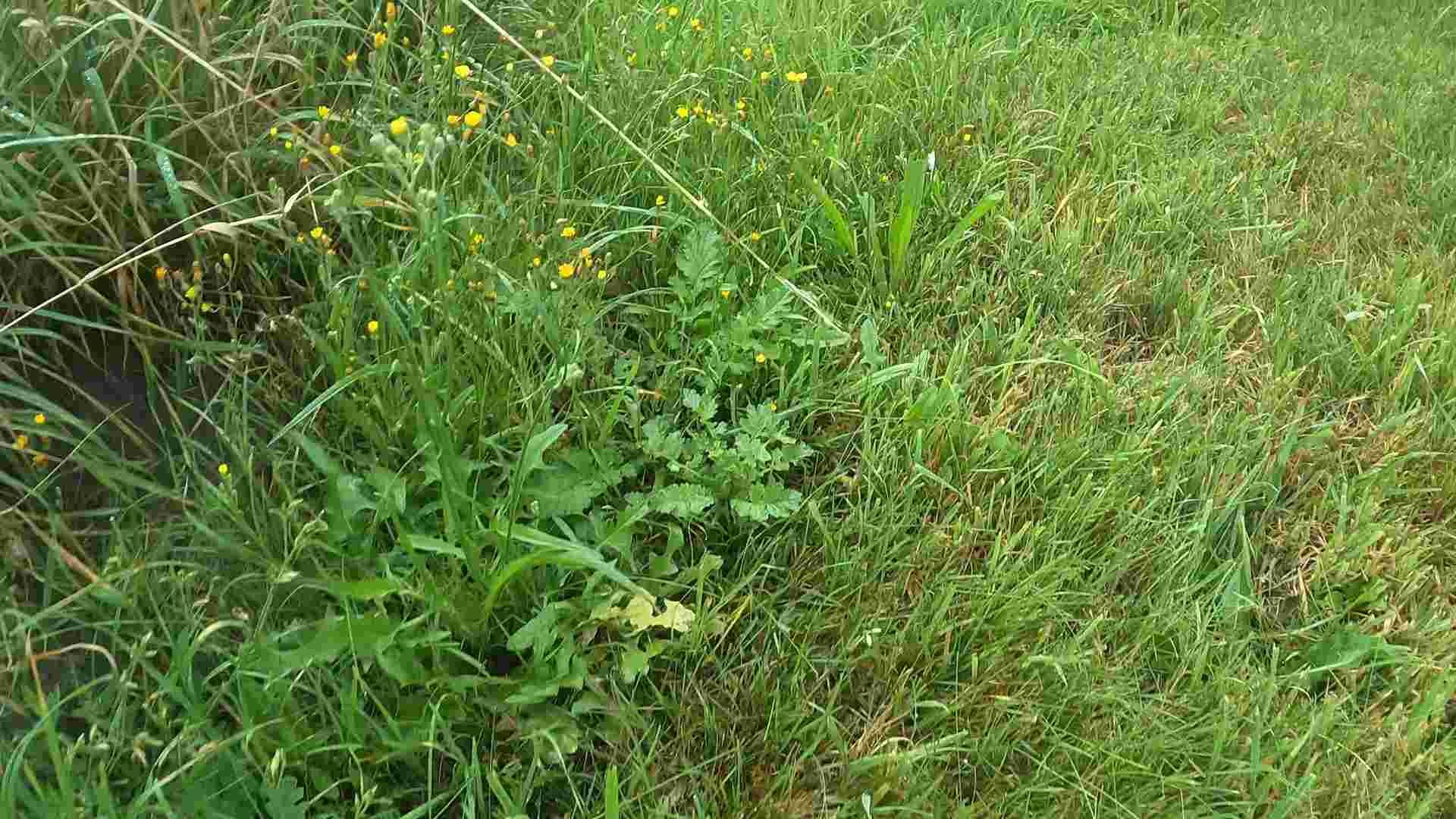}
        \caption{$\color{blue} \bullet$ D3}
    \end{subfigure}
    \hfill
    \begin{subfigure}[t]{0.22\linewidth}
        \includegraphics[width=\linewidth]{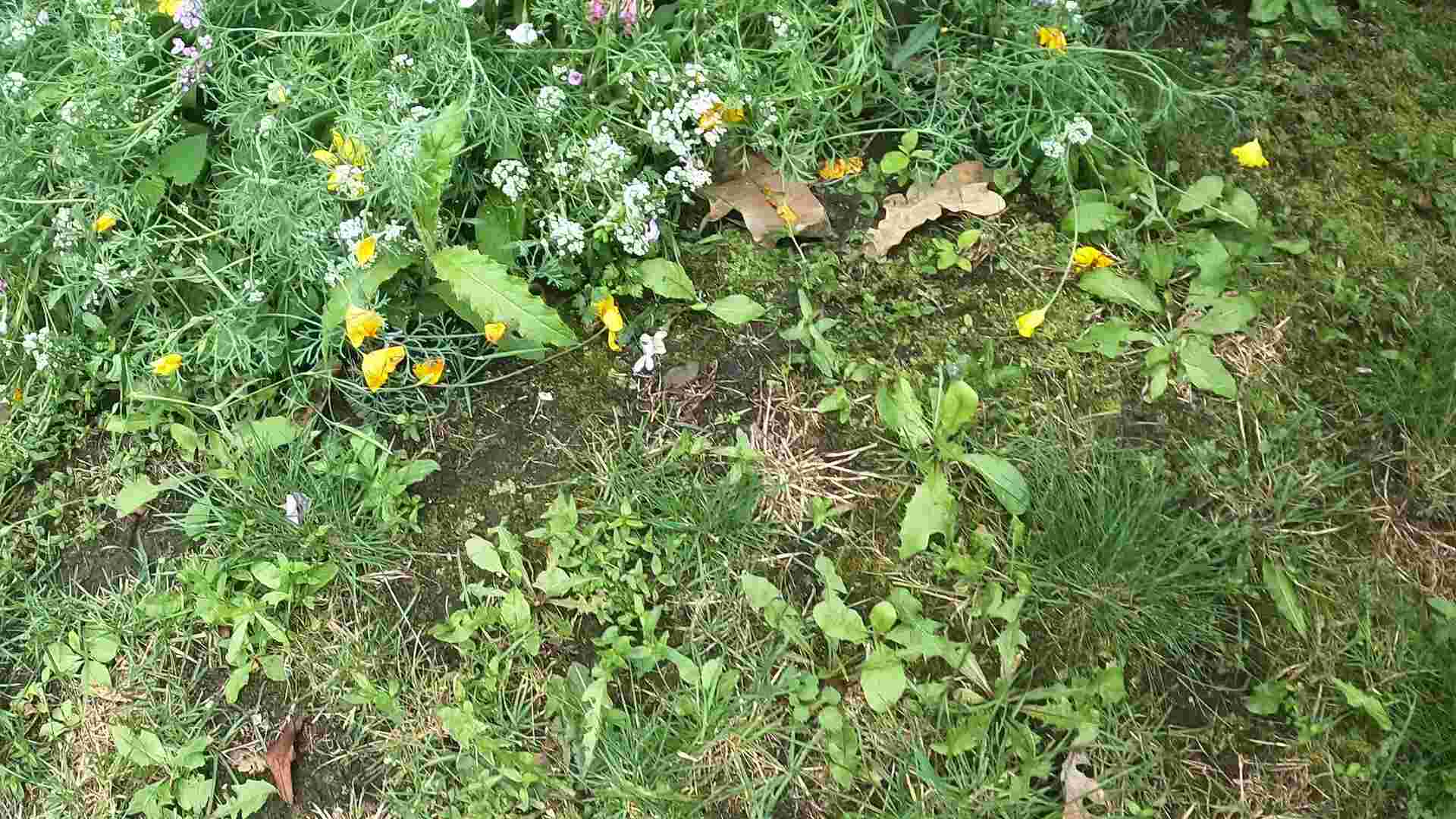}
        \caption{$\color{blue} \bullet$ D3}
    \end{subfigure}

    \vspace{2mm}

        \begin{subfigure}[t]{0.22\linewidth}
        \includegraphics[width=\linewidth]{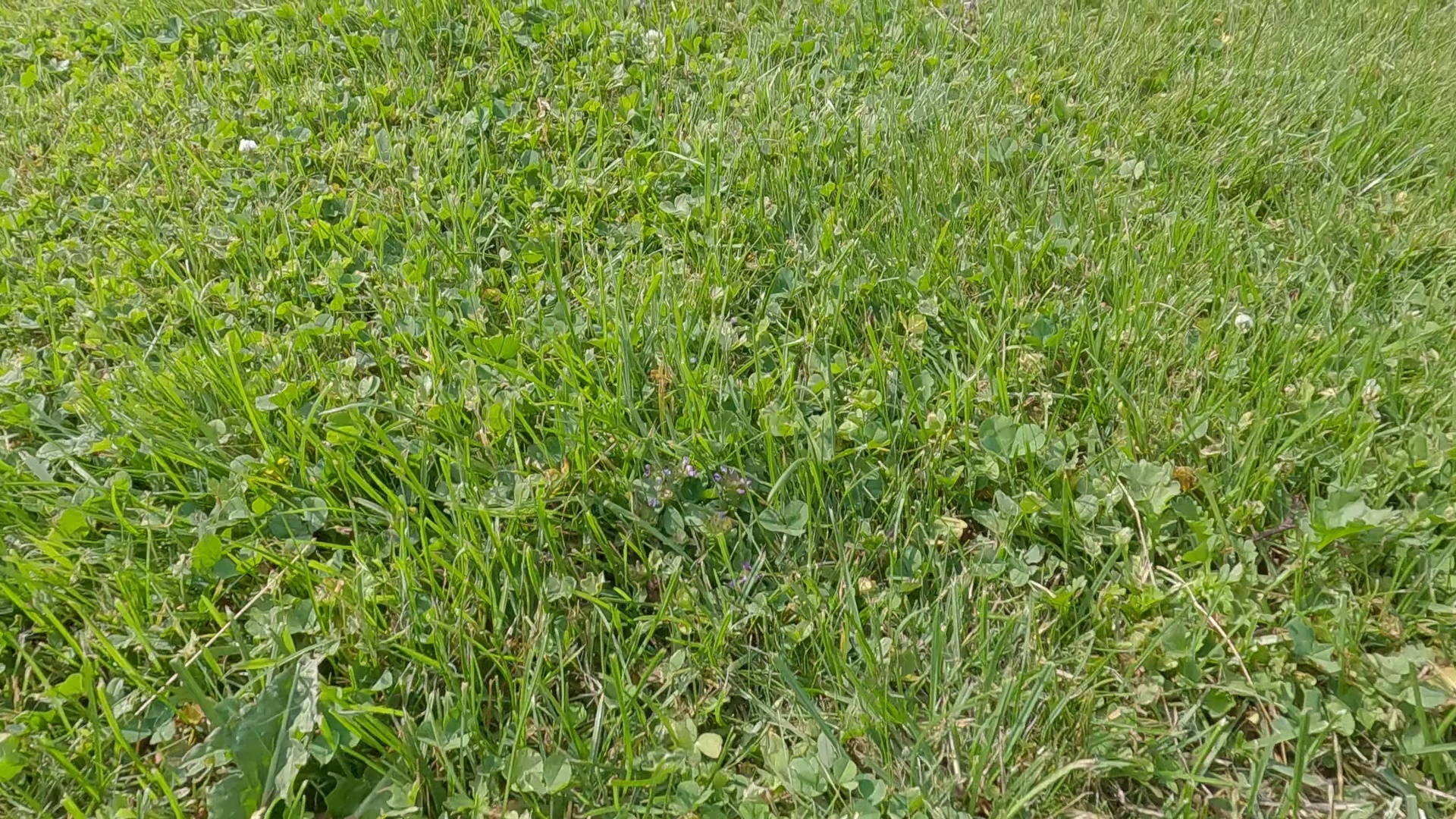}
        \caption{$\color{orange} \bullet$ D4}
    \end{subfigure}
    \hfill
    \begin{subfigure}[t]{0.22\linewidth}
        \includegraphics[width=\linewidth]{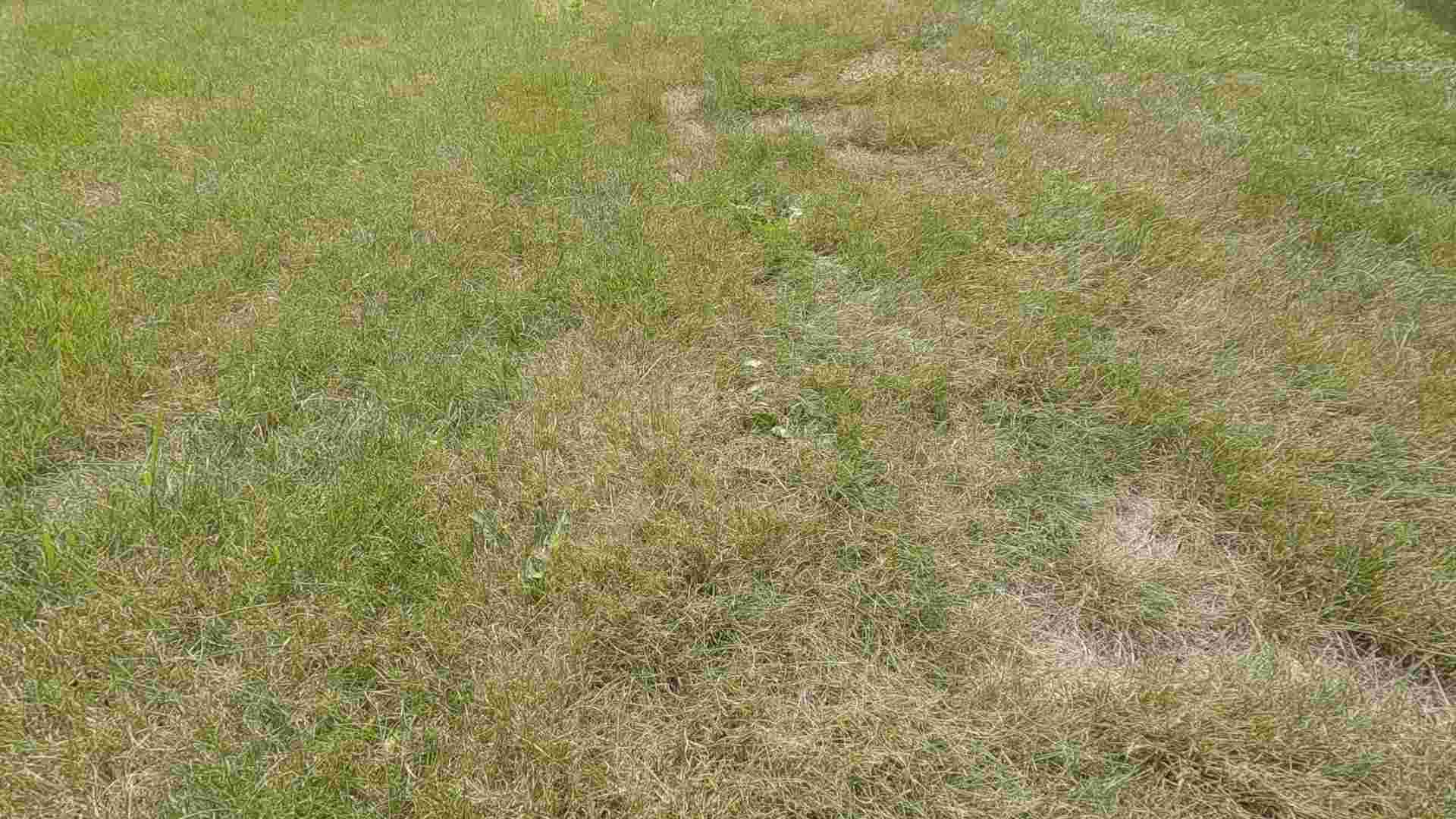}
        \caption{$\color{Green} \bullet$ D5}
    \end{subfigure}
    \hfill
        \begin{subfigure}[t]{0.22\linewidth}
        \includegraphics[width=\linewidth]{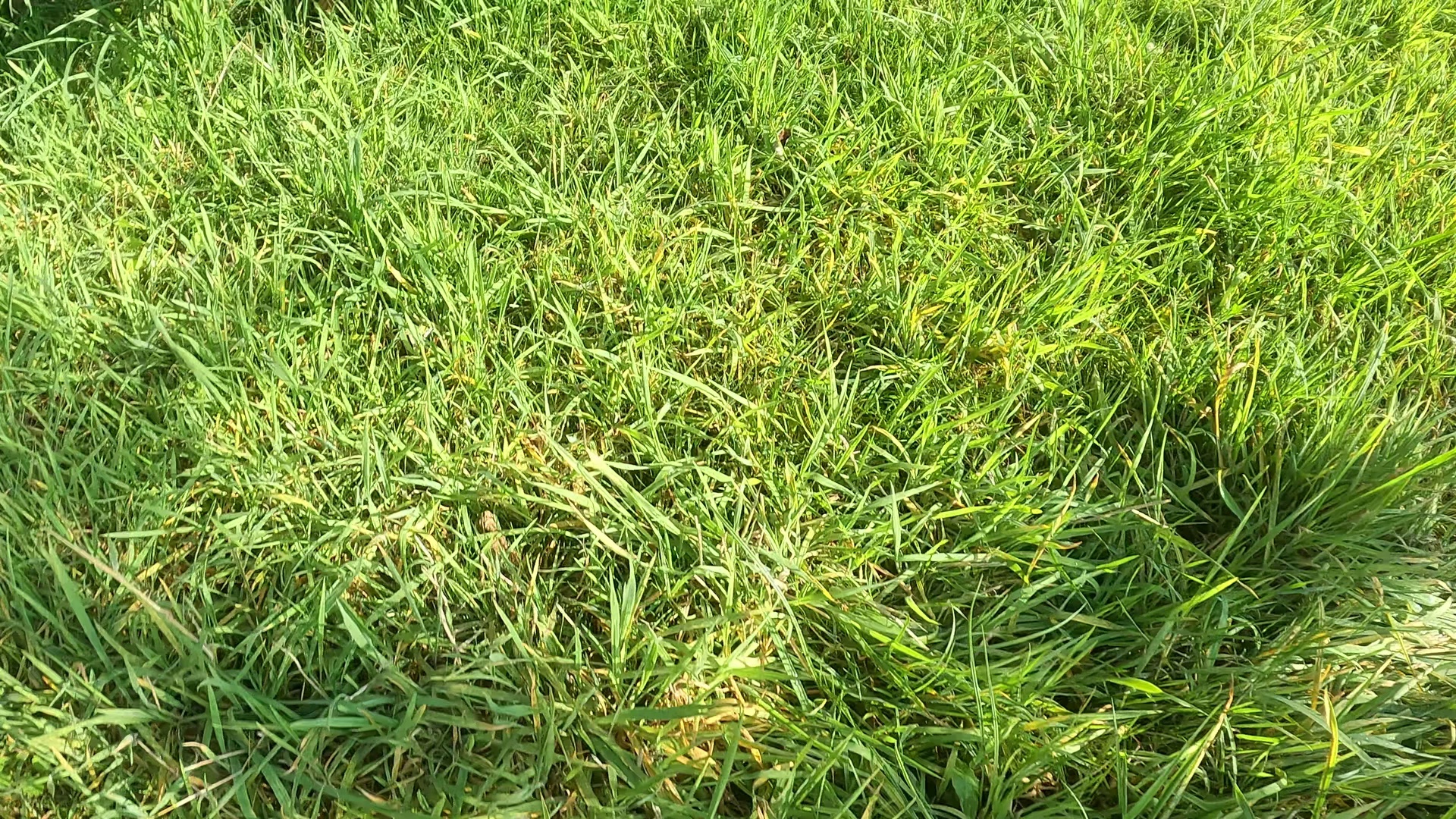}
        \caption{$\color{cyan} \bullet$ D6 }
    \end{subfigure}
    \hfill
    \begin{subfigure}[t]{0.22\linewidth}
        \includegraphics[width=\linewidth]{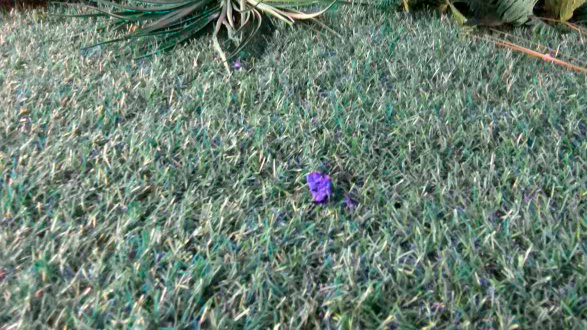}
        \caption{Mockup Lawn}
    \end{subfigure}

    \caption{Example frames from the collected lawn datasets, illustrating variation in biodiversity and mowing conditions. 
    (a)-(d): datasets with large biodiversity. (e)-(g): datasets with less biodiversity. (h): Image taken with our robot during the mock-up lawn experiment. These datasets are publicly available at \url{https://github.com/Lars-Beckers/BioBot}.}
    \label{fig:datasets_examples}
\end{figure*}

\begin{table*}[htb]
\centering
\caption{Overview of collected lawn image datasets used. Right: their calculated and manually assessed biodiversity scores.}
\label{tab:datasets}
\begin{tabular}{l l l | l  l }
\toprule
\textbf{Dataset ID} & \textbf{Description} & \textbf{Conditions} & \textbf{$\sigma_D$} & \textbf{Expert Score (1-5)}\\
\midrule
$\color{red} \bullet$ D1 &  Biodiverse lawn with green plants & Morning sunlight & 10.33 & 4\\
$\color{OliveGreen} \bullet$ D2 & Unmown biodiverse garden & Lightly cloudy  & 11.42 & 5\\
$\color{blue} \bullet$ D3 & Biodiverse flower-rich lawn & Bright sunlight & 12.56 & 5  \\
$\color{orange} \bullet$ D4 & Mown grass-only garden & Overcast & 8.70 & 1 \\
$\color{Green} \bullet$ D5 & Intensively mown lawn & Lightly cloudy & 9.13 & 2 \\
$\color{cyan} \bullet$ D6 & Long-grass garden & Strong sunlight & 9.51 & 2\\
\bottomrule
\end{tabular}
\end{table*}

To validate the biodiversity estimation method, a diverse set of lawn image datasets was collected under real outdoor conditions. 
All data were acquired using a Logitech BRIO webcam mounted on the robotic platform (see Fig.~\ref{fig:robot}), 
recording videos at a resolution of 1920$\times$1080 pixels. 
Video sequences were captured while the robot traversed various private gardens and test plots under different vegetation and lighting conditions.

From each recording, still frames were extracted at fixed temporal intervals to ensure spatial coverage without excessive redundancy. 
Frames were rescaled to $455\times256$ pixels and subsequently center-cropped to $224\times224$ pixels 
to match the input requirements of the ResNet50 backbone used for embedding extraction.  
The final dataset collection encompasses six distinct subsets, grouped according to their ecological characteristics and mowing status, as described in table~\ref{tab:datasets} and illustrated in figure~\ref{fig:datasets_examples}.

In total, 20840 frames were extracted, spanning a wide range of vegetation structures and lighting conditions. 
These datasets provide empirical data for evaluating the robustness of deep-feature biodiversity estimation 
under realistic, small-scale garden environments.

\subsection{Biodiversity estimation with deep embeddings}
\label{sec:embedding_space}

\subsubsection{Choice of the pre-training dataset}
A critical design choice in this work is the selection of an appropriate visual backbone for feature embedding.  
Since the goal is to estimate biodiversity directly from vegetation imagery, the pretrained model should already encode 
representations that are sensitive to botanical variation.  
For this reason, we adopted a ResNet50 network pretrained on the PlantNet300K dataset \cite{garcin2021plantnet}, 
a large-scale plant species classification corpus containing over 300,000 images across 1,081 species and 
captured under heterogeneous field conditions.  
This dataset was specifically designed for fine-grained plant recognition and thus offers a strong inductive bias toward 
texture, color, and morphological cues that are ecologically meaningful. Using a backbone pretrained on general-purpose datasets such as ImageNet would primarily emphasize 
object and shape recognition but not subtle intra-species variations between plant morphologies.  

\subsubsection{Visualizing the Embedding Space}

\begin{figure}[tb]
    \centering
    \includegraphics[width=1\linewidth]{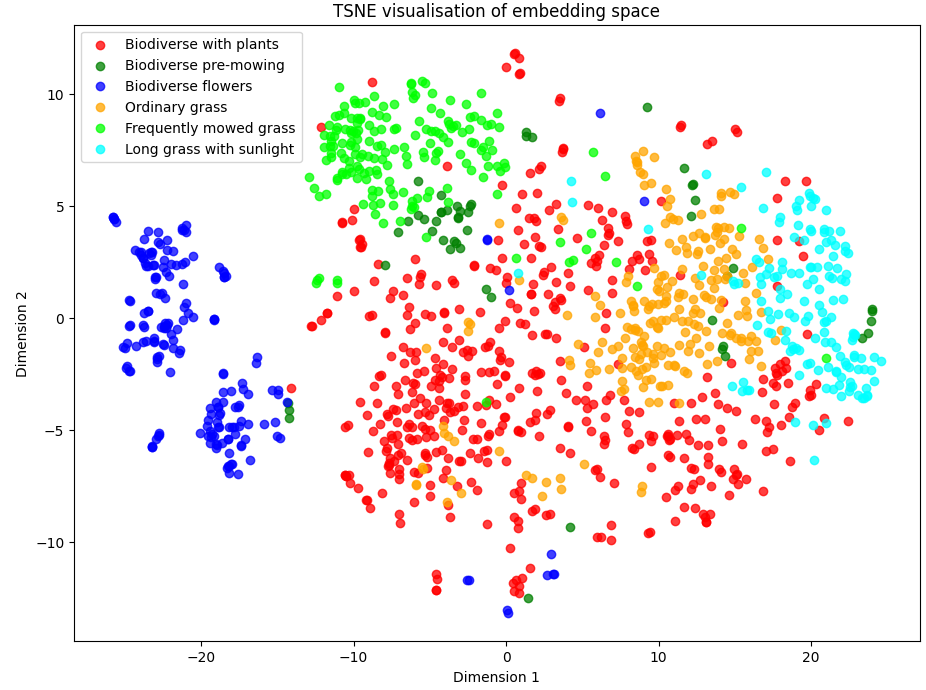}
    \caption{t-SNE visualization of embeddings from all biodiverse (red, blue) and non-biodiverse (orange, cyan) datasets 
    using the PlantNet300K-pretrained ResNet50 model.  
    Greater spatial spread indicates higher visual diversity within the dataset.}
    \label{fig:tsne}
\end{figure}

To explore how the pretrained model organizes real garden imagery, we processed all datasets of table~\ref{tab:datasets} and projected their embeddings onto a two-dimensional manifold using t-SNE 
(Fig.~\ref{fig:tsne}).    
This visualization provides qualitative insight into how the network perceives visual differences between varying vegetation conditions.
A clear separation is observed among datasets: biodiverse gardens (in red and dark blue) occupy a broad and dispersed region in the embedding space, indicating high visual heterogeneity.  
In contrast, the homogeneous grass datasets (orange) cluster more tightly, reflecting lower intra-class variability.  
Interestingly, the dataset containing flowering plants (blue) is positioned further away from the main cluster, consistent with its distinct chromatic and structural features.  

A subtler but important pattern is seen between two grass-dominated datasets (orange and cyan): although both mainly contain grass species, their overlap in embedding space is limited.  
This suggests that the embeddings are influenced not only by the semantic visual structure of the vegetation, 
but also by contextual and environmental parameters such as illumination, moisture, 
or soil coloration. These influences reinforce the necessity of the initial pre-mowing patrol stage we propose, during which the robot samples the specific range of local visual and lighting conditions before any mowing takes place.

The clear separation between datasets in Fig.~\ref{fig:tsne} validates that the pretrained ResNet50 model provides a meaningful representation space in which ecological and visual diversity can be quantified geometrically. This observation justifies its use as the foundation for computing global deviation and local density-based biodiversity measures in the following analyses.

To compare the biodiversity between datasets, we computed a global descriptor based on the dispersion of image embeddings in feature space.  Each dataset was first processed by the ResNet50 backbone to extract deep feature vectors 
$\{ f_i \in \mathbb{R}^d \}$ from the penultimate layer for every image.  
The global deviation $\sigma_D$ of a dataset $D$ was then defined as the mean Euclidean 
distance of all feature vectors to the global embedding-space centroid $\bar{f}$:

\[
\sigma_D = \frac{1}{N} \sum_{i=1}^{N} \| f_i - \bar{f} \|_2,
\qquad \text{with } 
\bar{f} = \frac{1}{N} \sum_{i=1}^{N} f_i.
\]

This metric captures the overall spread of vegetation representations in the learned feature space.  
Datasets with diverse vegetation (e.g., containing flowering plants, weeds, or mixed grasses) 
tend to produce embeddings that are more dispersed, resulting in a higher $\sigma_D$.  
Conversely, visually uniform lawns (e.g., recently mown or monocultural grass) yield lower deviation values.

To assess ecological validity, a horticultural expert independently assigned a qualitative biodiversity score 
(1 = very low diversity; 5 = very high diversity) to each dataset based on manual visual inspection in the field.  
The righthand part of table~\ref{tab:datasets} summarizes the computed global deviation values $\sigma_D$ and the corresponding expert assessments.

% \begin{table}[h]
% \centering
% \caption{Global deviation values (average Euclidean distance to embedding-space centre) and expert biodiversity scores for different datasets.}
% \label{tab:globaldeviation}
% \begin{tabular}{lccc}
% \toprule
% \textbf{Dataset} & \textbf{Category} & \textbf{Global Deviation $\sigma_D$} & \textbf{Expert Score (1–5)} \\
% \midrule
% D1 & Biodiverse lawn 1 (plants) & 11.42 & 5 \\
% D2 & Biodiverse lawn 2 (flowers) & 12.56 & 5 \\
% D3 & Biodiverse lawn 3 (unmown garden) & 10.33 & 4 \\
% D4 & Mown grass lawn & 9.51 & 2 \\
% D5 & Intensively mown lawn & 8.70 & 1 \\
% D7 & Mown lawn 3 & 9.13 & 2 \\
% \bottomrule
% \end{tabular}
% \end{table}

% %version without column
% \begin{table}[h]
% \centering
% \caption{Global deviation values (average Euclidean distance to embedding-space centre) and expert biodiversity scores for different datasets.}
% \label{tab:globaldeviation}
% \begin{tabular}{lccc}
% \toprule
% \textbf{Dataset} & \textbf{Category} & \textbf{Global Deviation $\sigma_D$} & \textbf{Expert Score (1–5)} \\
% \midrule
% D1 & Biodiverse lawn 1 (plants) & 11.42 & 5 \\
% D2 & Biodiverse lawn 2 (flowers) & 12.56 & 5 \\
% D3 & Biodiverse lawn 3 (unmown garden) & 10.33 & 4 \\
% D4 & Mown grass lawn & 9.51 & 2 \\
% D5 & Intensively mown lawn & 8.70 & 1 \\
% D7 & Mown lawn 3 & 9.13 & 2 \\
% \bottomrule
% \end{tabular}
% \end{table}

A clear correlation is observed: datasets visually identified as more biodiverse exhibit larger global deviations in the embedding space.  
This confirms that the proposed representation-space dispersion metric effectively reflects ecological richness, 
supporting its use as a label-free proxy for biodiversity assessment in garden-scale environments.

\subsection{Biodiversity-Aware Mowing Experiments}

To evaluate the complete perception–action loop of the proposed system, we conducted controlled mowing experiments 
using both a mock-up test field and real-world lawn imagery.  
The experiments aimed to verify whether the robot’s decision algorithm could reliably identify visually diverse vegetation patches 
and adapt its mowing behavior accordingly.

\subsubsection{Mock-up Lawn Experiment}

\begin{figure}[tb]
    \centering
    \includegraphics[width=0.9\linewidth]{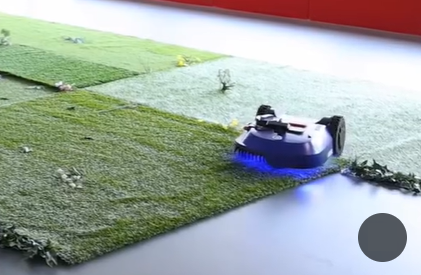}
    \includegraphics[height=0.8\linewidth]{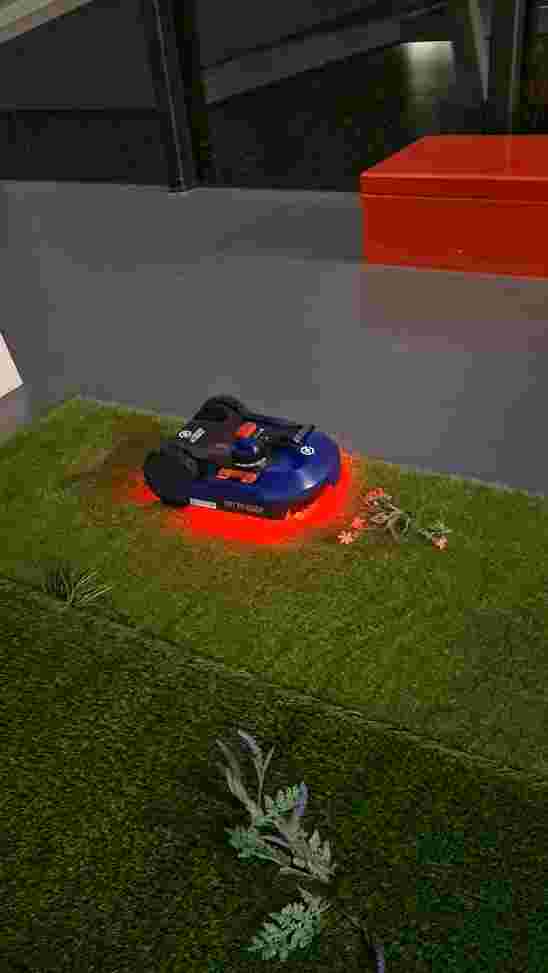}
    \includegraphics[height=0.8\linewidth]{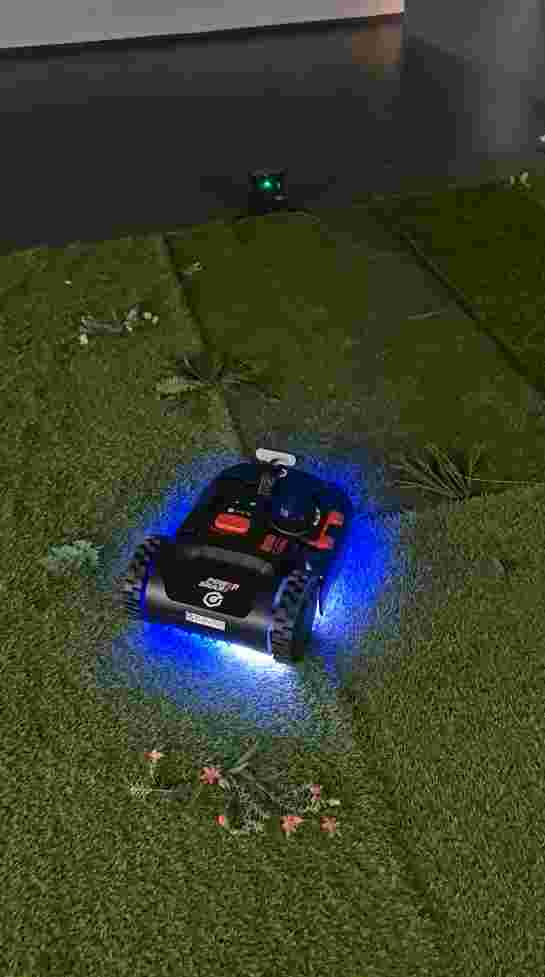}
    \caption{Mock-up lawn experiment. A video can be seen at \url{https://www.youtube.com/shorts/n4Pn7bTxB2s}}
    \label{fig:mockup_demo}
\end{figure}

A $6m\times4m$ mock-up lawn was constructed indoors using artificial grass as a background.  Imitation flowering plants and plastic weeds were randomly distributed across the surface to simulate local biodiversity patches.  This setup, shown in Figure~\ref{fig:mockup_demo}, provided full control over lighting and scene composition while allowing repeatable testing without affecting real vegetation. Figure~\ref{fig:datasets_examples} (h) shows an example frame collected during this experiment.

The robotic mower prototype was placed at one edge of the mock-up field and executed the complete procedure described in Section~\ref{sec:procedure}.  To provide immediate visual feedback, a strip of RGB LEDs mounted underneath the robot indicated the decision state. 

Through iterative testing, the decision threshold $\tau$ and neighborhood parameter $k$ were fine-tuned 
until the robot consistently activated mowing over grass-only regions and deactivated it near the flowers.  
The final tuned system successfully demonstrated selective mowing behavior: flowers and diverse patches were preserved, while uniform grass was “mown”.

\subsubsection{Evaluation on Real Lawn Data}

After validating the behavior on the mock-up field, the same decision algorithm and tuned parameters 
were applied to real lawn imagery from the collected datasets.  
The algorithm processed each frame independently, generating binary mowing decisions 
based on the global embedding-space statistics derived from the patrol phase.

\begin{figure}[tbp]
    \centering
    \includegraphics[width=0.4\columnwidth]{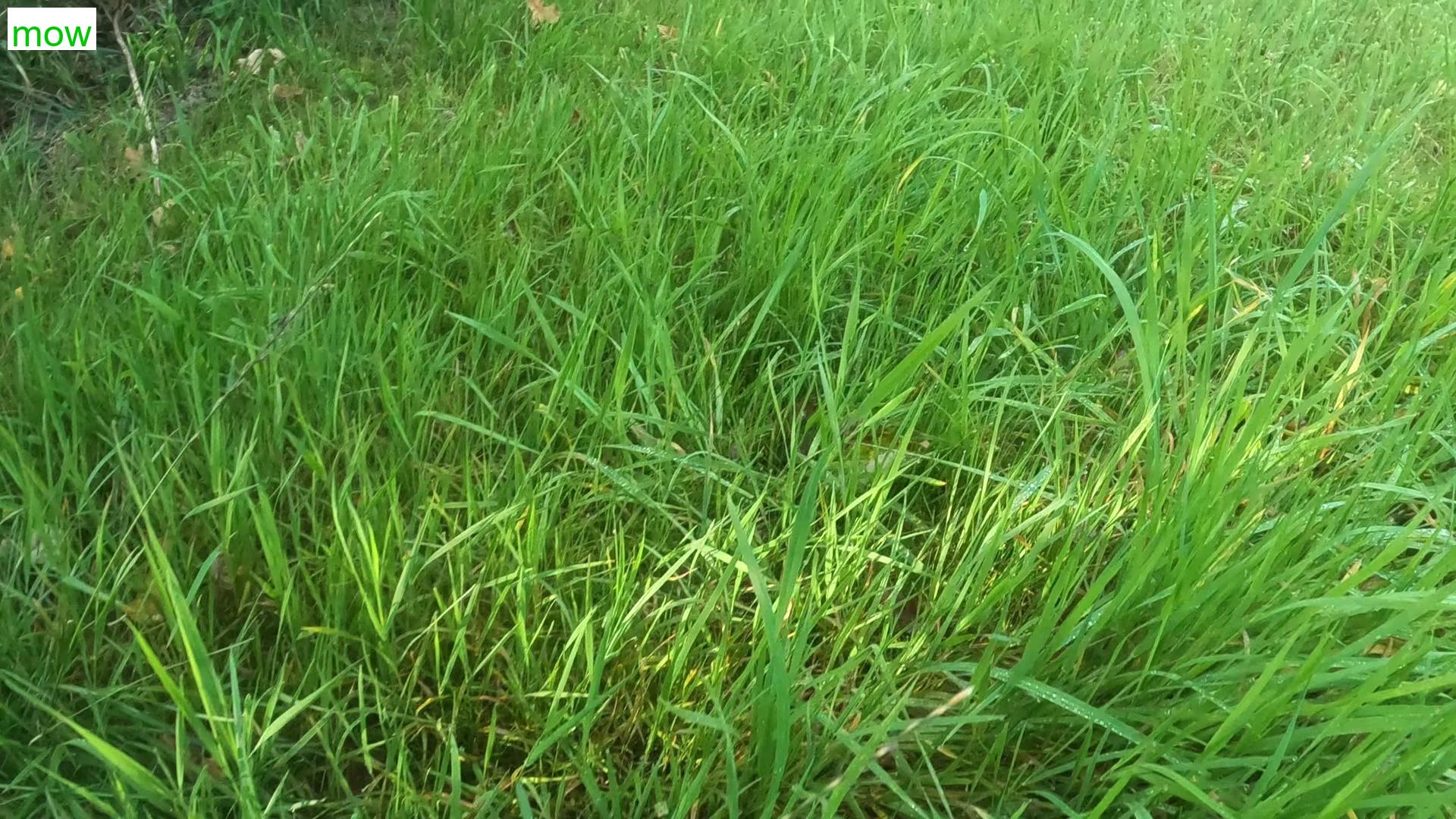}
    \includegraphics[width=0.4\columnwidth]{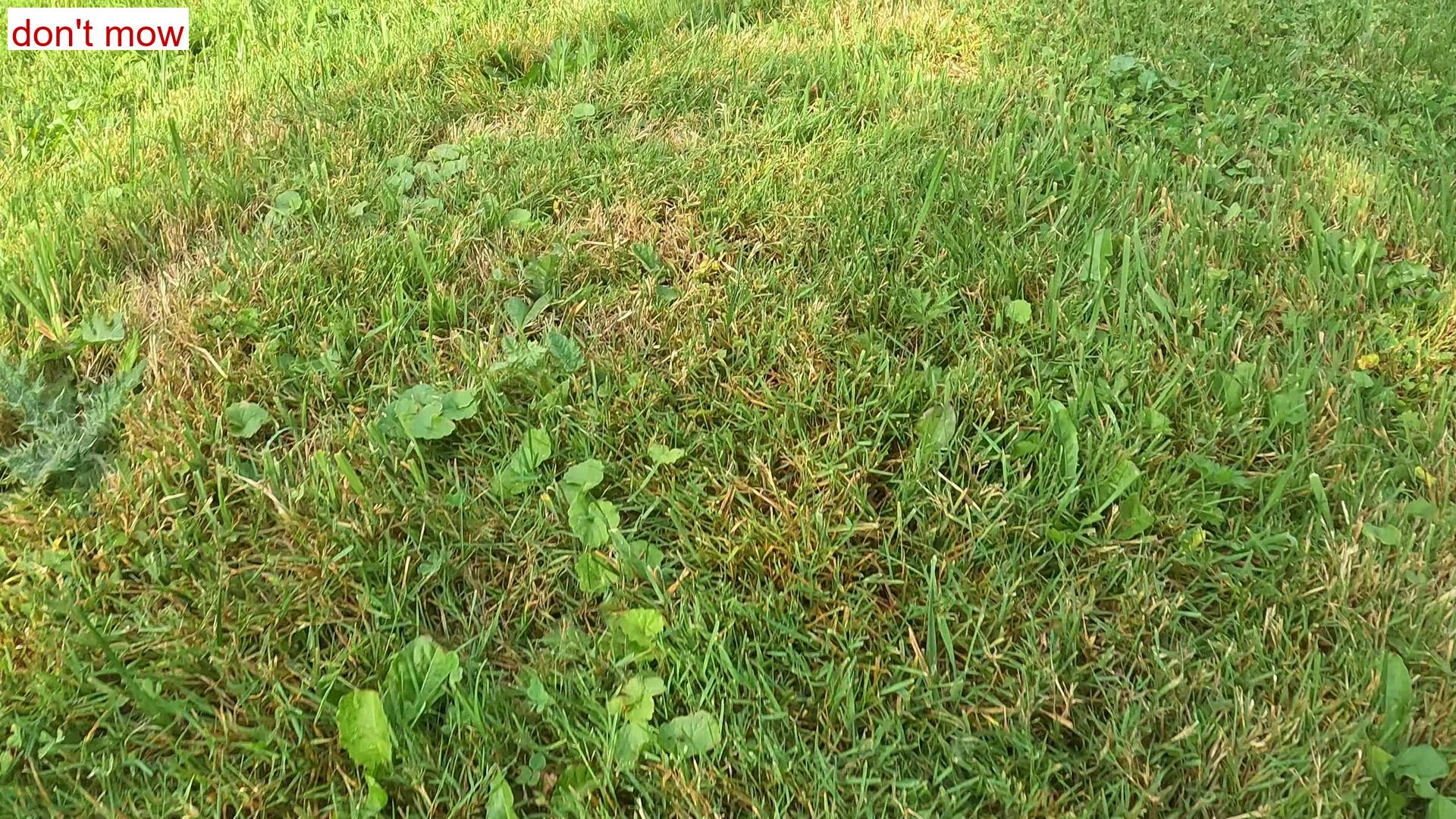}
    \includegraphics[width=0.4\columnwidth]{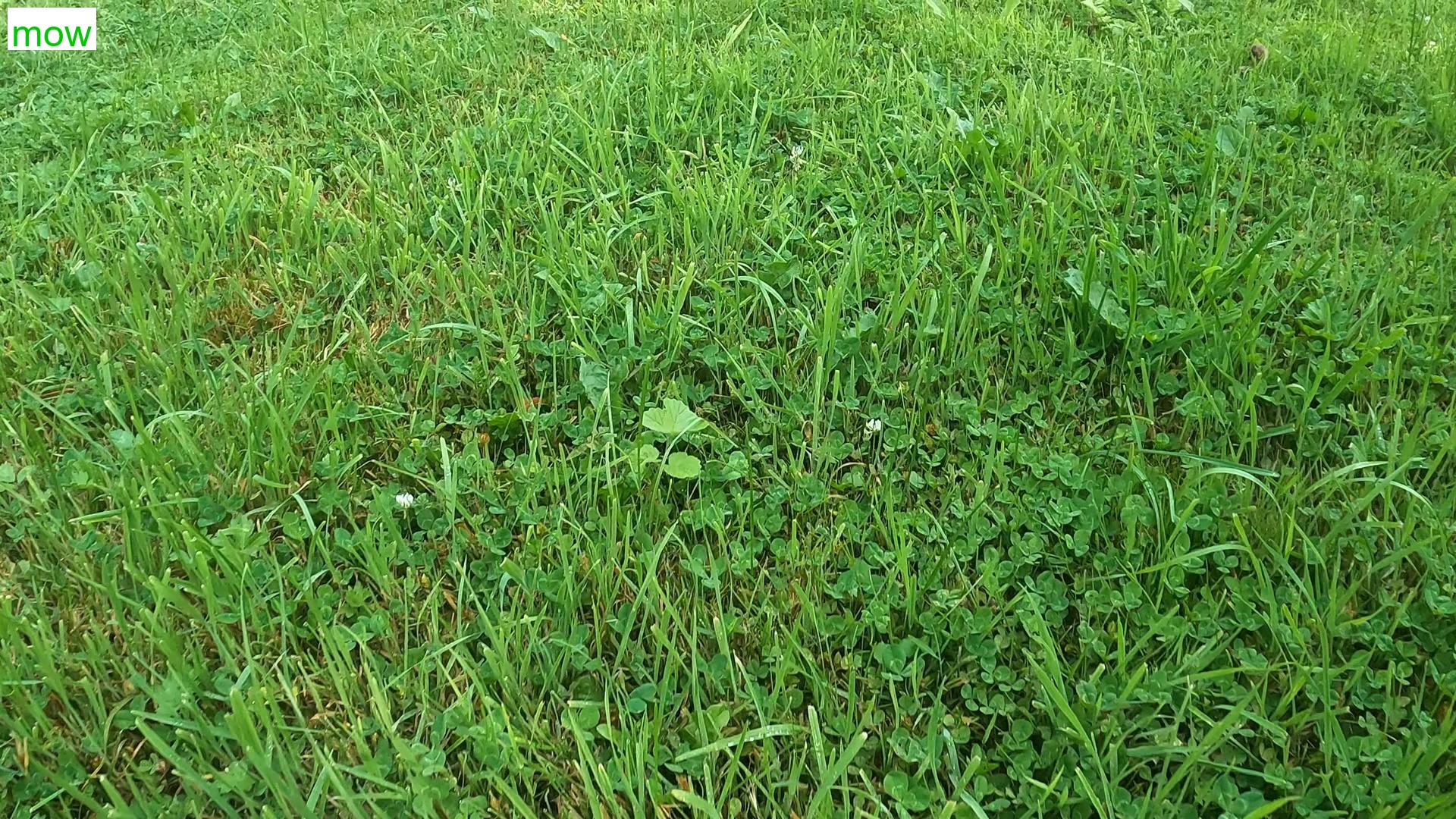}
    \includegraphics[width=0.4\columnwidth]{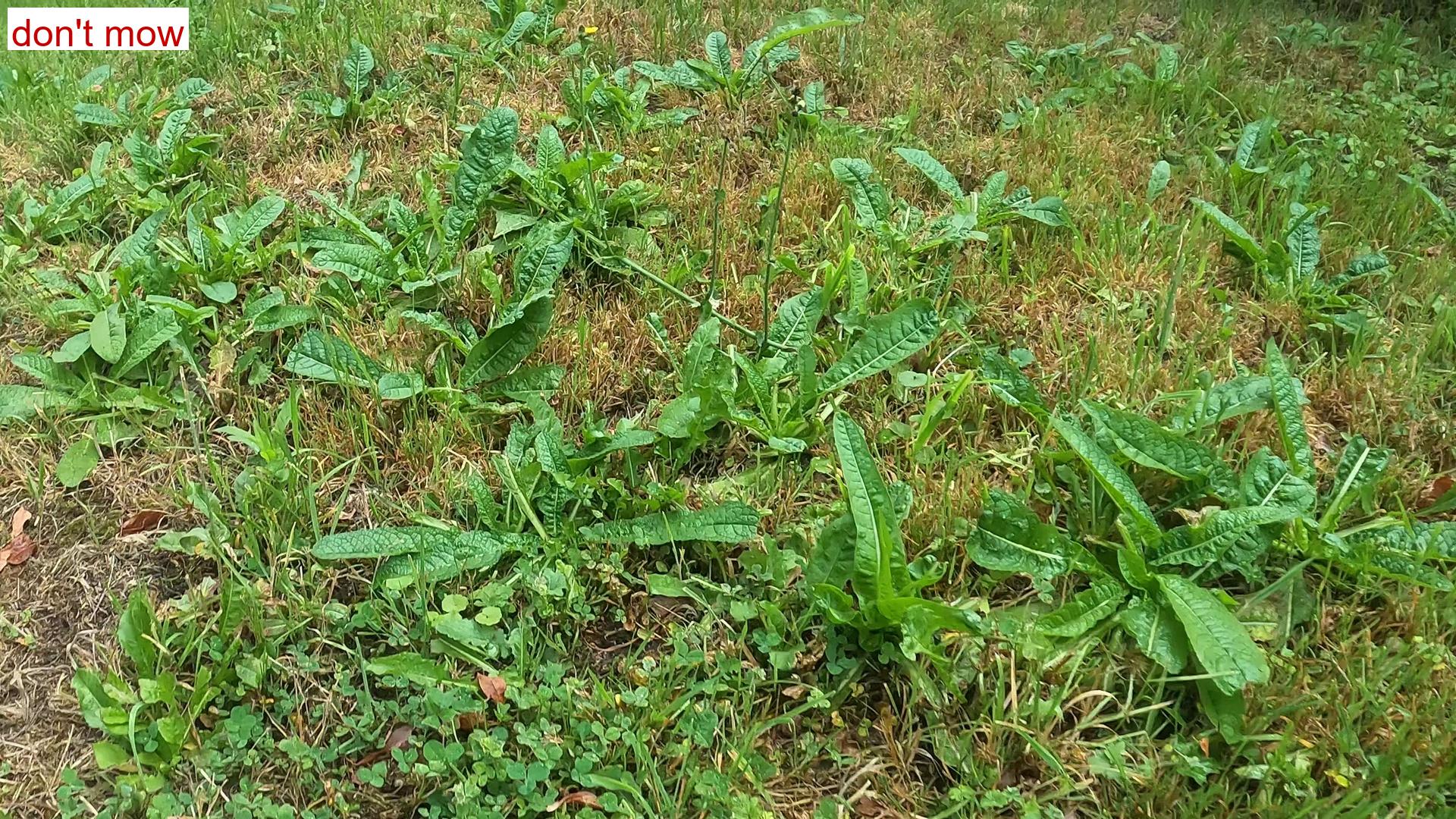}
    \includegraphics[width=0.4\columnwidth]{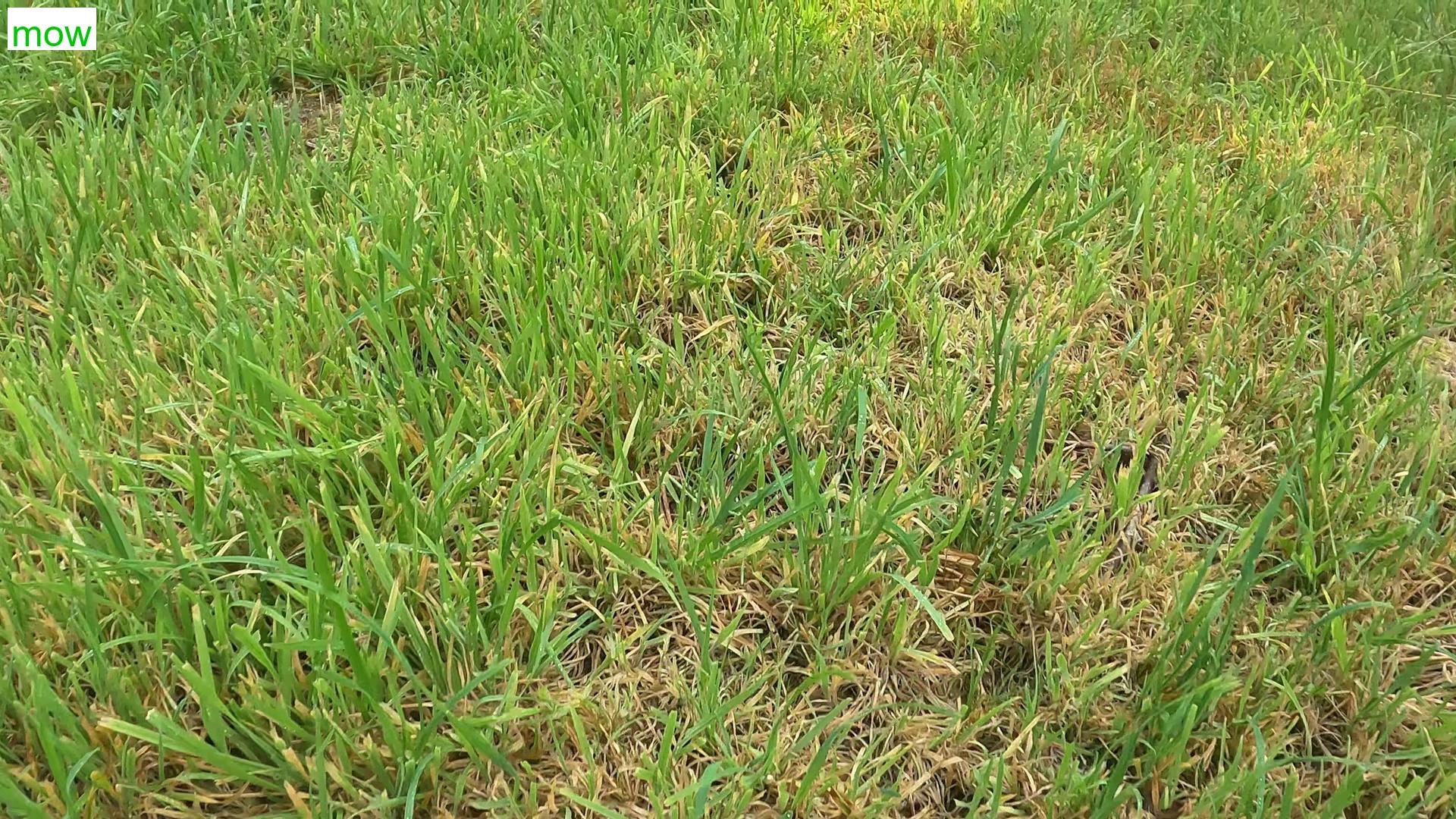}
    \includegraphics[width=0.4\columnwidth]{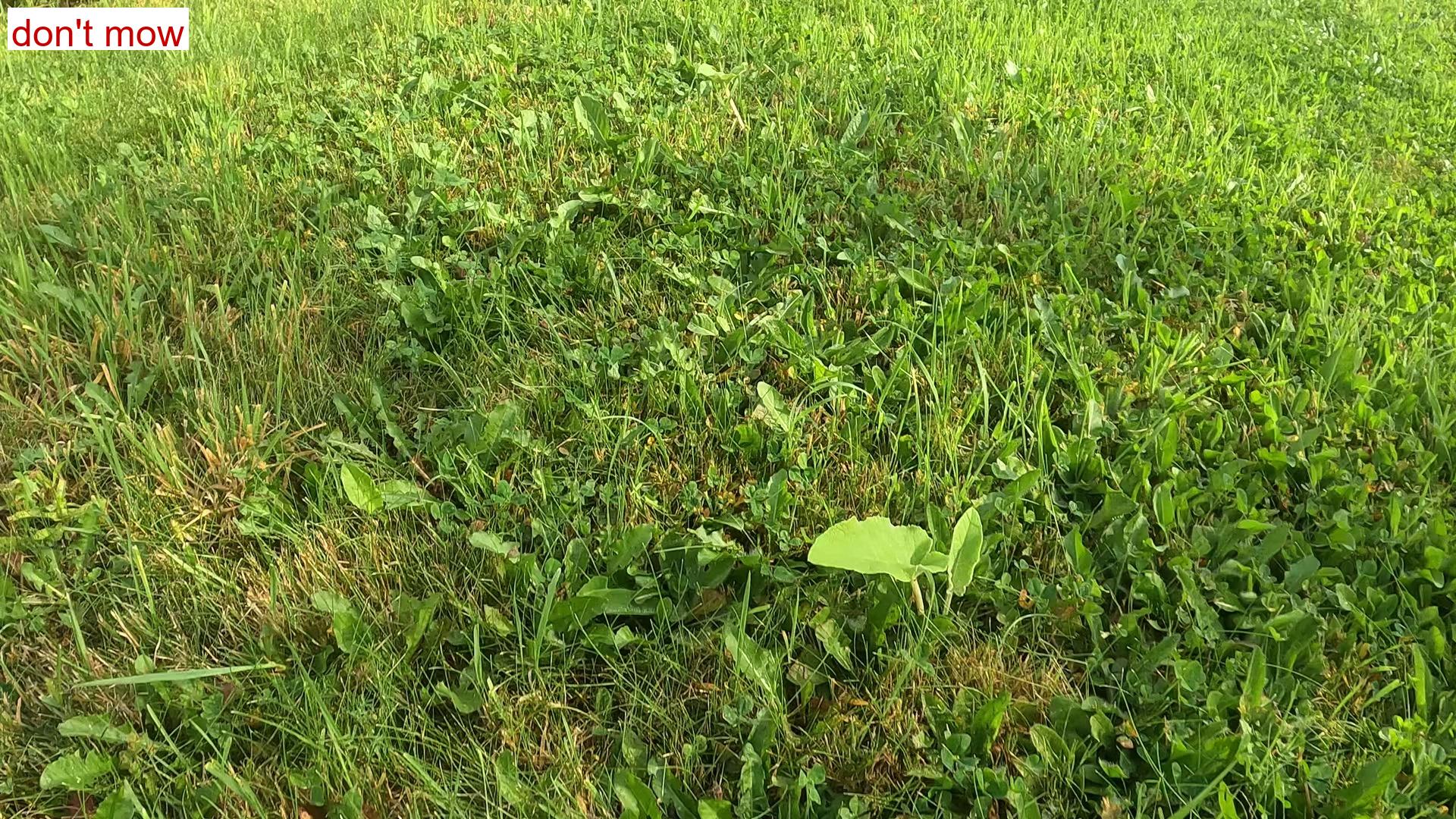}
    \includegraphics[width=0.4\columnwidth]{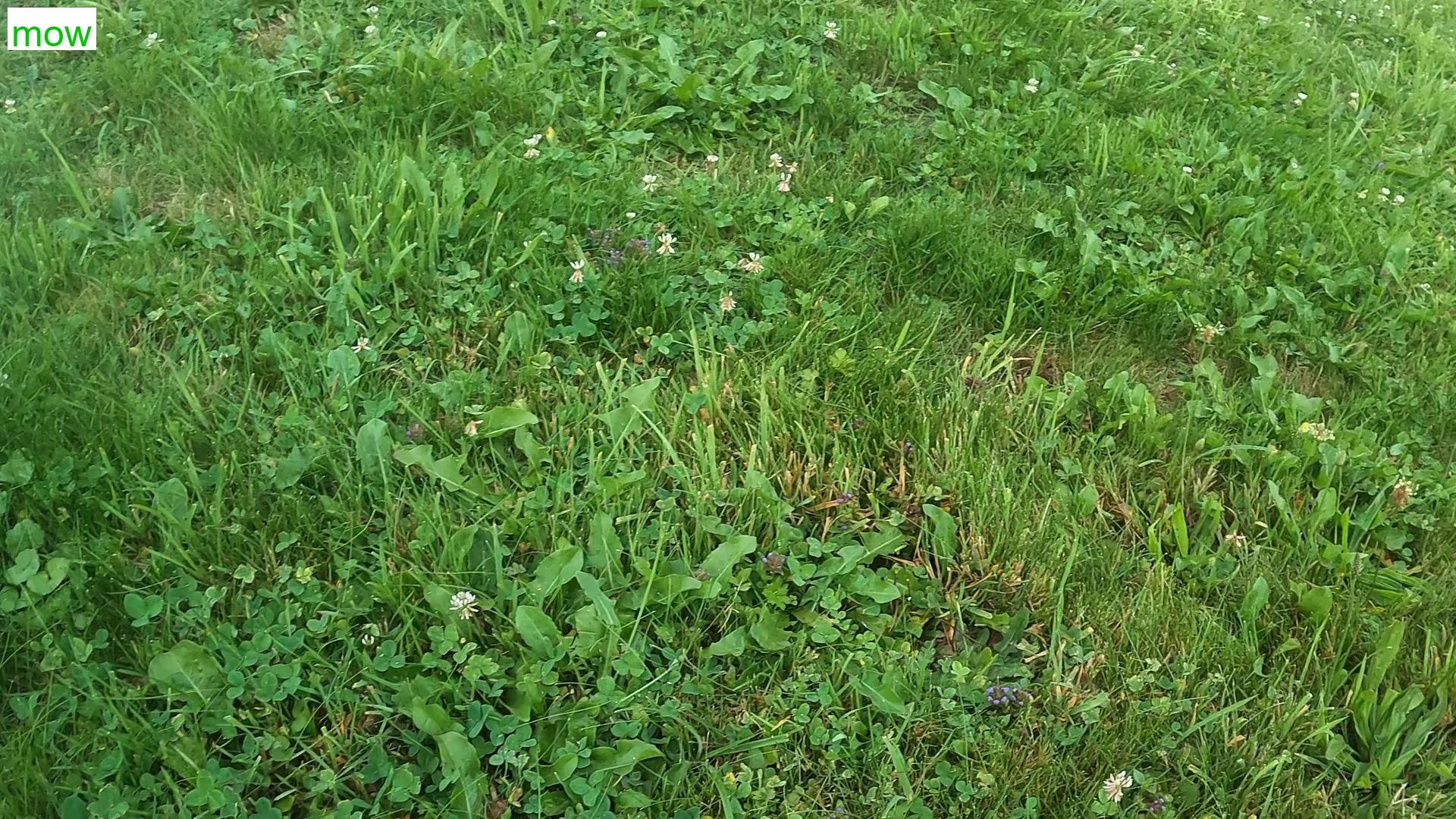}
    \includegraphics[width=0.4\columnwidth]{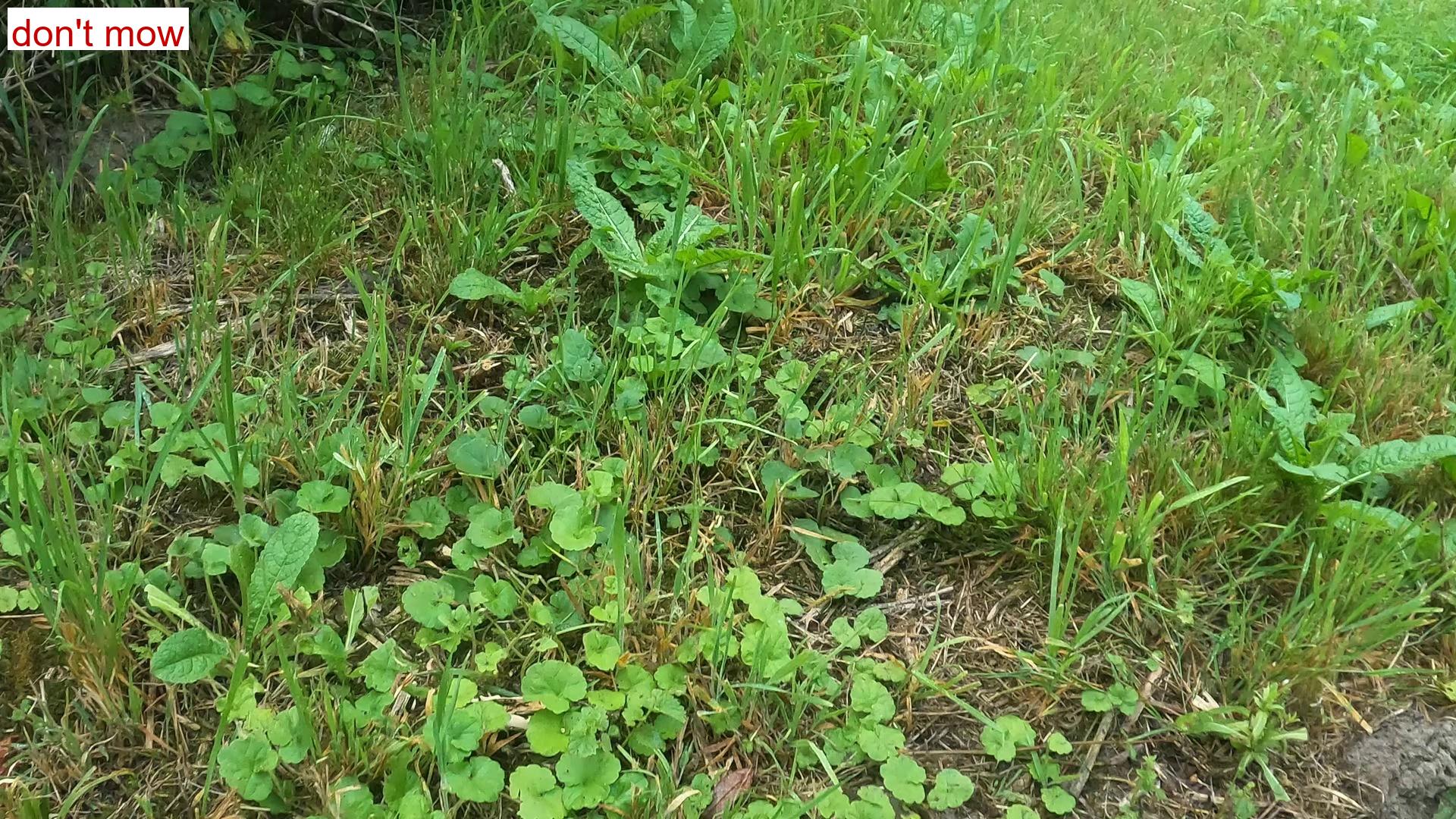}
     
    \caption{Example frames from the real-lawn dataset processed by the biodiversity-aware mowing algorithm.  
    (Left) Frames classified as “mow”; (Right) frames classified as “don't mow.” \\Full video available at: \url{https://youtu.be/fEt5faNmKG8}}
    \label{fig:reallawn_results}
\end{figure}

Qualitatively, the results were consistent with the mock-up tests:  
frames dominated by homogeneous grass were flagged as mowable, while images containing mixed vegetation were marked as non-mowable.  
Figure~\ref{fig:reallawn_results} presents example frames from the real-lawn evaluation, 
showing clear visual differentiation between mowing and conservation areas.

These results confirm that the proposed vision-based biodiversity estimation approach can drive practical mowing decisions both in controlled and natural environments. The embedding-space representation, pre-calibrated during the patrol phase, 
successfully transfers to unseen vegetation conditions, demonstrating the feasibility of computer vision for fine-grained, biodiversity-aware robotic behavior.

\section{CONCLUSION AND FUTURE WORK}

This paper presented a novel approach to autonomous lawn maintenance that explicitly integrates biodiversity awareness into the control loop of a robotic mower.  We developed a functional demonstrator equipped with camera-based perception and real-time 
mowing control, enabling the robot to estimate biodiversity directly from visual data.  Using a ResNet50 model pretrained on the PlantNet300K dataset, we extracted deep embeddings that capture ecologically meaningful variations in plant morphology and color.  
A global deviation metric based on the average Euclidean distance to the embedding-space centre was proposed as a simple, label-free proxy for biodiversity estimation. Experiments across multiple lawn datasets confirmed a clear correlation between the embedding dispersion and expert-assessed biodiversity scores.

To close the perception–action loop, we implemented an adaptive mowing decision algorithm that modulates blade activation based on local visual diversity.  The system was first validated on a mock-up lawn, where imitation flowers were consistently spared while grass patches were “mown.” When applied to real lawn imagery, the same algorithm demonstrated consistent differentiation between uniform and biodiverse vegetation regions. These experiments confirm that the proposed pipeline can operate as an effective proof of concept for biodiversity-increasing robotic mowing.

If adopted widely at the household level, biodiversity-increasing mowing robots could collectively transform private gardens—one of the largest green surface areas in densely populated regions—into a distributed ecological network that strengthens urban biodiversity.

While the presented results demonstrate technical feasibility, first future work will be longer-term ecological field validation campaigns with real flora, necessary to quantify long-term biodiversity outcomes.  In addition, integrating a probabilitsic mowing approach or adding a sensing modality for estimating vegetation height
would enable the system to detect and control overgrowth, ensuring that some patches do not remain perpetually unmown and turn into dense “jungle” areas.

\section*{ACKNOWLEDGEMENTS}
This work is partially supported by the Flanders AI Research initiative of the Flemish Government.

\bibliographystyle{apalike}
{\small
\bibliography{references}}

\end{document}